\documentclass[letterpaper, 10 pt, conference]{ieeeconf}  % Comment this line out if you need a4paper

\UseRawInputEncoding

\usepackage{graphicx} % for pdf, bitmapped graphics files
\usepackage{times} % assumes new font selection scheme installed
\usepackage{amsmath} % assumes amsmath package installed
\usepackage{amssymb}  % assumes amsmath package installed
\usepackage{url}
\usepackage{enumerate}
\usepackage{multirow}
\usepackage{colortbl}
\usepackage{booktabs} % for top and bottom line
\usepackage{algorithmicx}
\usepackage{algorithm}
\usepackage{makecell}
\usepackage{algpseudocode}% http://ctan.org/pkg/algorithmicx
\usepackage{cite}
\usepackage{bm}

\makeatletter
\let\NAT@parse\undefined
\makeatother
\usepackage[hidelinks]{hyperref} %hide the borders of the cross-references

\begin{document}

\title{\LARGE \bf
MMFN: Multi-Modal-Fusion-Net for End-to-End Driving 
\author{Qingwen Zhang$^{1}$, Mingkai Tang$^{1}$, Ruoyu Geng$^{1}$, Feiyi Chen$^{1}$, Ren Xin$^{1}$, Lujia Wang$^{1,2}$}

\thanks{$^{1}$Authors are with Robotics Institute, The Hong Kong University of Science and Technology, Hong Kong SAR, China. \texttt {\{qzhangcb, mtangag, rgengaa, fchenak, rxin\}}@connect.ust.hk}
\thanks{$^{2}$Lujia Wang is the corresponding author and also with Clear Water Bay Institute of Autonomous Driving. \texttt {eewanglj}@ust.hk}

% \thanks{$^{1}$Qingwen Zhang, Ruoyu Geng, and Ren Xin are with The Hong Kong University of Science and Technology (Guangzhou), Nansha, Guangzhou, 511400, Guangdong, China. \texttt \{qzhangcb, rgengaa, rxin\}@connect.ust.hk}
% \thanks{$^{2}$Mingkai Tang, Feiyi Chen, Lujia Wang are with The Hong Kong University of Science and Technology, Clear Water Bay, Hong Kong SAR, China \texttt \{mtangag, fchenak\}@connect.ust.hk; Lujia Wang is the corresponding author and also with Clear Water Bay Institute of Autonomous Driving. eewanglj@ust.hk}
}
\maketitle

%%%%%%%%%%%%%%%%%%%%%%%%%%%%%%%%%%%%%%%%%%%%%%%%%%%%%%%%%%%%%%%%%%%%%%%%%%%%%%%%
\begin{abstract}
% 摘要整体思路：A 介绍传感器引入初衷，B 说明hard to use in end-to-end driving，C 指出我们如何efficient use, and fusion，并得出好的结果
Inspired by the fact that humans use diverse sensory organs to perceive the world, sensors with different modalities are deployed in end-to-end driving to obtain the global context of the 3D scene. 
In previous works, camera and LiDAR inputs are fused through transformers for better driving performance. These inputs are normally further interpreted as high-level map information to assist navigation tasks. Nevertheless, extracting useful information from the complex map input is challenging, for redundant information may mislead the agent and negatively affect driving performance. 
We propose a novel approach to efficiently extract features from vectorized High-Definition (HD) maps and utilize them in the end-to-end driving tasks.
In addition, we design a new expert to further enhance the model performance by considering multi-road rules. Experimental results prove that both of the proposed improvements enable our agent to achieve superior performance compared with other methods. The source code is released as an open-source package.
\end{abstract}
%%%%%%%%%%%%%%%%%%%%%%%%%%%%%%%%%%%%%%%%%%%%%%%%%%%%%%%%%%%%%%%%%%%%%%%%%%%%%%%%

\section{INTRODUCTION}

% importance of end-to-end driving & limitation learning 
Autonomous driving is conducted via several modules\cite{gog2021pylot,liu2020hercules}, namely localization\cite{jiao2021greedy}, perception\cite{liu2021ground}, planning\cite{cheng2022real}, and control. However, the system performance is constrained by the manually selected intermediate criteria, e.g., localization and lane detection error. 
One solution is to use end-to-end driving which optimizes the system from the perspective of the overall system performance, avoiding the potential loss caused by incorrect human-designed intermediate criteria. 
In this work, we apply imitation learning to train an agent to mimic the expert's action and behavior with the available sensor data inputs.

\begin{figure}[t]
\centering
\includegraphics[width=3.3in]{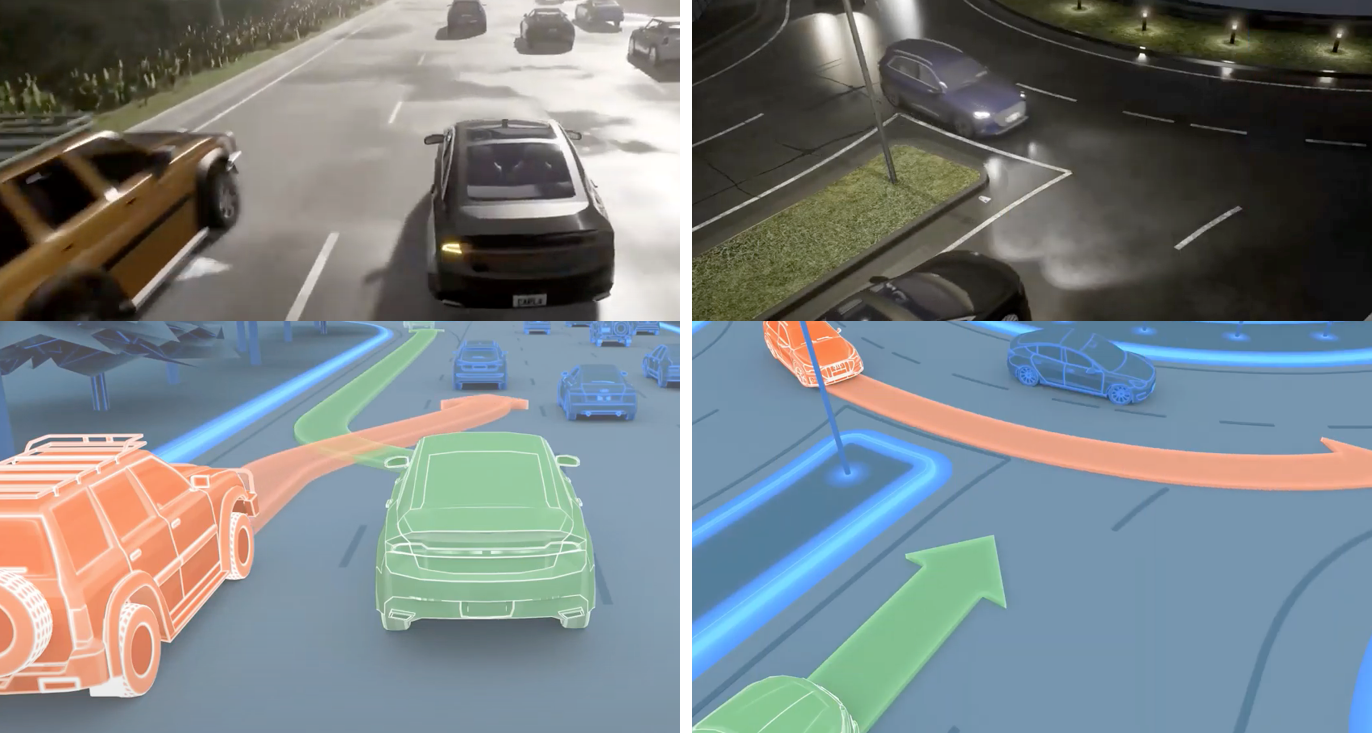}
\caption{Scenarios in complex urban environments. The ego vehicle needs to yield to other vehicles when switching lanes or at the roundabout. These scenarios indicate more abundant environmental information such as the lane structure and the velocities of other vehicles may assist to better facilitate the end-to-end driving task. Figures captured from \href{https://www.youtube.com/watch?v=-L9VuPpzVdQ}{CARLA leaderboard}.}
\label{fig:backgroud}
\vspace{-1.5em} % for figure gap smaller
\end{figure}

% expert 解释
In imitation learning, expert drivers are utilized to collect the training data and generate its ground truth of actions before the training phase. Experts in  \cite{lbc, transfuser, neat} only consider avoidance collisions according to the distance between agents. As illustrated in the left image of Fig. \ref{fig:backgroud}, it is dangerous to only consider the nearby vehicles in terms of distance, since vehicles may be at a high speed when changing lanes. Hence, more information is added in our expert to enable the expert to capture more potential collisions which are missed in existing methods.

% motivation of sensor fusion
Apart from the ground truth given by the expert, the agent in imitation learning needs sensor data to have a clear understanding of the surroundings. Existing methods in end-to-end driving mostly depend on only a single type of sensor, like camera\cite{lbc} or LiDAR\cite{cai2021carl}. 
Recently, Prakash \textit{et al.} proposed to fuse camera and LiDAR data through an attention mechanism in \cite{transfuser}. It turns out that taking advantage of the complementary sensors can achieve a satisfactory result. Specifically, cameras can better capture texture information. However, they are susceptible to lighting conditions. LiDAR outperforms cameras in terms of accurate distance information, whereas the sparsity of LiDAR point clouds may cause information loss. Apart from the intrinsic disadvantages of these two sensors, driving scenarios in complicated urban environments, examples of which as illustrated in Fig. \ref{fig:backgroud}, also reveal the importance of introducing more sensors into end-to-end driving. 

% \footnotetext[3]{}

The scenarios presented in Fig. \ref{fig:backgroud} suggest the benefits of providing lane structures and vehicle velocities for the end-to-end driving task. Thus, we add the HD map and radar on top of the LiDAR and camera as the network inputs in our approach. The HD map offers high-level map data like lanes and roads, which is more closely related to the action or trajectory outputs compared with camera and LiDAR data. Additionally, radar can offer the velocities of other agents directly without any calculation, which are more accurate than the ones calculated from other sensor data. However, little work has been conducted on how to efficiently represent HD maps and radar data in end-to-end learning frameworks and extract useful features from them. It is challenging to effectively integrate the complex HD map, which contains not only hierarchical geometric information but also semantic information, into the network. When integrating radar data into the network, the sparsity of radar points can not be ignored.
Therefore, we propose a Multi-Modal-Fusion-Network (MMFN) to properly represent and fuse the meaningful information extracted from the complementary sensor raw data in the end-to-end driving task. Our method is proved to be effective in CARLA \cite{carla} leaderboard task and open-sourced in \color{blue}\href{https://github.com/Kin-Zhang/mmfn}{https://github.com/Kin-Zhang/mmfn}. \color{black}The main contributions of this work include the following:
\begin{itemize}
    \item We use a multi-path consideration rule-based expert to improve the performance of existing agents \cite{lbc}.
    \item We explore different representations on the HD map as the network input and propose a framework based on \cite{transfuser} to fuse the four different types of sensor data. %Specifically, the vectorized map considerably improves the driving performance.
    % 本地和在线测试
    \item We experimentally validate the performance of our method both on local driving routes and the online leaderboard\cite{leaderboard}, with an increase in the driving score by 34.67\% and a decrease in the infraction rate by 50.8\% on average compared with \cite{transfuser}.
\end{itemize}

\section{RELATED WORK}
In end-to-end driving, many works only concentrate on utilizing a single type of sensor, like cameras. For example, Chekroun \textit{et al.}~\cite{GRI} used three cameras and combined the action result and the semantic segmentation result to train the agent jointly. Similarly, Chitta \textit{et al.}~\cite{neat} also employed three cameras to get neural attention fields and predicted the future waypoints. Other works like \cite{wor} and \cite{marl} both focused on camera input and used deep reinforcement learning to circumvent the use of an expert. These works show that applying multiple networks \cite{resnet,CNN,yolov3} initially proposed for image detection can promote the driving performance \cite{NVIDIA-2016}. As mentioned before, cameras are not robust to lighting conditions, so information resources such as LiDAR, HD maps, and radar, have been added to compensate for the limitations of cameras.

% different methods to process different sensor data
To use other sensors apart from cameras, processing techniques tailored for each sensor need to be utilized. Images like RGB and RGBD data with semantic labels are processed similarly using off-the-shelf networks like convolutional neural networks (CNNs). In the object detection task using LiDAR, Lang \textit{et al.}~\cite{pointpillars} proposed to extend the dimensions of point data to offer more detailed information. Rhinehart \textit{et al.}~\cite{rhinehart2019precog} presented a compact form to describe the LiDAR point cloud data, and this was utilized in our method. Specifically, the LiDAR point clouds are divided into two groups based on their z-axis values and a pseudo image with the number of points inside each pixel of the image is output. Gao \textit{et al.}~\cite{vectornet} represented the map in a vectorized form when predicting other agents' motion. While Chen \textit{et al.} \cite{lbc} used a bird's-eye-view (BEV) map to express the map, which included more information, like the location of other agents and traffic lights, compared with \cite{vectornet}.

% existing works on multi-modality fusion networks
Most existing multi-modality fusion networks focus on perception tasks like object detection and motion forecasting. For instance, Liang \textit{et al.}~\cite{liang2019multi} designed a network for 3D object detection to jointly process camera and LiDAR inputs, of which the outputs are the results of four sub-tasks. Liang \textit{et al.}~\cite{liang2018deep} and Guo \textit{et al.}~\cite{guo2021deep} fused the multi-modal features obtained from BEV LiDAR and RGB images in a multi-scale fusion fashion for 3D object detection. Their work inspired other researchers to introduce the sensor fusion mechanism into end-to-end driving. For example, Prakash \textit{et al.}~\cite{transfuser} proposed to embed LiDAR point clouds into a transformer network to cooperate with camera image data. Based on \cite{transfuser}, we additionally integrate HD map and radar data into the network. We also show the effective use of the transformer network by applying different types of sensors and their corresponding data representations.

\section{METHODOLOGY}
\begin{figure*}[ht]
\centering
\includegraphics[width=7in]{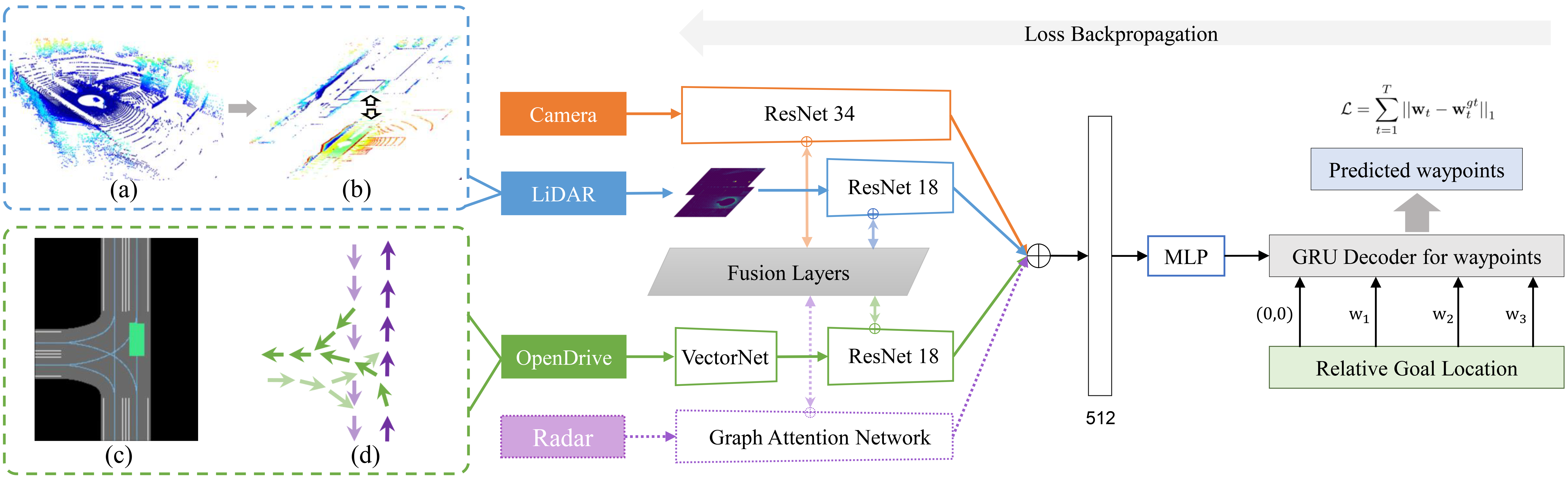}
\caption{\textbf{Architecture}. Process on LiDAR Point Cloud in Section~\ref{section:pc}. (a) receive all point cloud data, (b) divide point clouds with 2m height into two channels; Two methods for representing the HD map as input to the network in Section~\ref{section:map}: (c) rasterize to BEV perspective image, (d) vectorize road elements.}
\label{fig:arc}
\vspace{-1.0em} % for figure gap smaller
\end{figure*}

% TODO 是否需要添加一个problem formulation
In this section, we improve the expert in \cite{lbc} by considering time to collision and lane structures. Then, we directly use the dataset expert collected to train the MMFN framework which introduces multiple sensors. Fig. \ref{fig:arc} shows the whole framework for training, details of how to extract sensor data feature can be found in the following. 
% In this section, we explain the improvements in the expert, detail the representations of the four sensor inputs and the output, and introduce the critical network components, the feature extraction, and the fusion module.
\subsection{Expert Principle}
\begin{figure}
\centering
\includegraphics[width=3.0in]{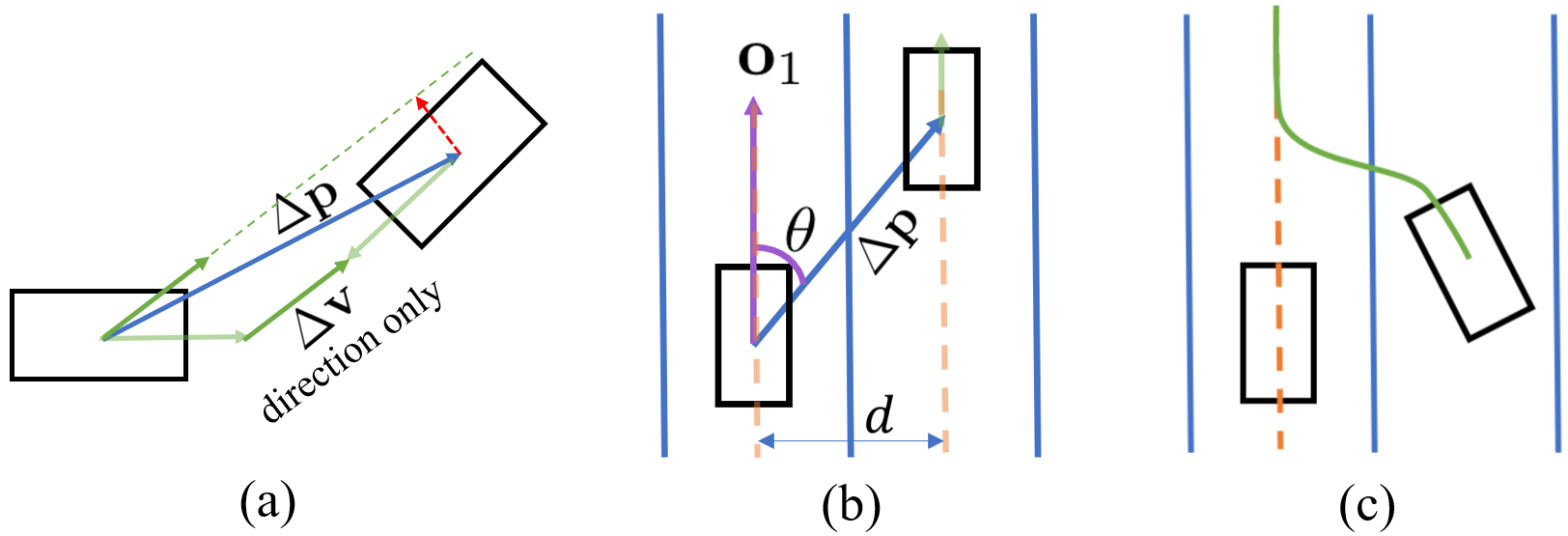}
\caption{Illustrations of expert rules under different scenarios. (a) the scenario when calculating TTC, and the red dashed line is $ \Delta \mathbf p - \mathbf{P}_{p2v} $; (b) the scenario when the ego vehicle and the nearby vehicle are driving on the same road but in different lanes, where $d$ is the current lane width obtained from the HD map; (c) the scenario when the ego car changes its lane but there is another car in the adjacent lane.}
\label{fig:expert}
\vspace{-1.0em} % for figure gap smaller
\end{figure}

% map information added to the expert
The expert agent in \cite{lbc} (AUTO expert) only uses distance to find the nearby obstacles and only considers its front area. This means that the AUTO expert fails to locate its nearby agents using the road or lane ID. To address this issue, the map information provided by the CARLA simulator\cite{carla} is used in our expert agent, where the map refers to the lane information attached to vehicles. Our expert can drive more similarly to human drivers, because the traffic rules that human drivers abide by are based on lane structures that can be obtained from the map input. Based on the AUTO expert, we add time to collision (TTC) and take the vehicles in other lanes into consideration when lane switching happens. 

% TTC
To calculate the TTC, the expert first gets the relative position vector $\Delta \mathbf p$ and relative velocity $\Delta \mathbf v$ between two cars in the CARLA simulator, and $\Delta \mathbf v$ is projected to $\Delta \mathbf p$ through:
\begin{align}
\label{figure_a}
\mathbf P_{v2p}=\Delta \mathbf p \cdot \frac{\Delta \mathbf p^\top \Delta \mathbf v}{\Delta \mathbf p^\top\Delta \mathbf p}, \quad \text{TTC}=\frac{\mathbf P_{v2p}}{\Delta \mathbf p},
\end{align}
where $\mathbf P_{v2p}$ is the projected relative velocity. Moreover, $\left \| \Delta \mathbf p - \mathbf{P}_{p2v} \right \| $, as illustrated in Fig. \ref{fig:expert} (a), is the unreachable distance. When it exceeds the threshold defined by the lane width $d$, the expert thinks that this nearby vehicle will not collide with the ego vehicle in terms of TTC. 

% STOP
For the vehicle stop case, not only is the distance considered but also the angle, and the angle from the ego vehicle to the obstacle is calculated by
\begin{align}
\label{figure_b}
\theta = \frac{\arccos{(\mathbf o_1 \cdot \Delta \mathbf p)}}{\left \| \mathbf o_1 \right \| \left \| \Delta \mathbf p \right \| },
\end{align}
where $\mathbf o_1$ corresponds to the orientation of the ego vehicle. When $\theta>\arcsin{\frac{d}{\left \| \Delta \mathbf p \right \|}}$, it means that the obstacle may not be in front of the ego car, as Fig. \ref{fig:expert} (b) indicates, where $d$ is the current lane width.

% CHANGE lane
As shown in Fig. \ref{fig:expert} (c), when the ego vehicle intends to switch to other lanes, apart from the locations of other vehicles, the lane information of these vehicles will also be extracted from the HD map. If another vehicle is in the lane that the ego vehicle intends to switch to, and the distance between these two vehicles is below a certain threshold, the ego vehicle will stop for a while and then execute changing.

\subsection{Input and Output Representation}
%introduce the newly added sensors
To make the best of the fusion mechanism in \cite{transfuser}, apart from camera and LiDAR data, an OpenDrive HD map and radar data are added as the network inputs in the proposed approach. Even though the HD map in this work is obtained from the CARLA simulator, there are still several approaches to export the HD map automatically \cite{AUTOHDmap, MULTIHDmap}. Additionally, to improve the model's adaptability to dynamic environments, long-range radar is used. In the following, we briefly introduce the processing techniques for each type of sensor data used in our method.

\subsubsection{OpenDRIVE HD Map}
\label{section:map}
Considering that maps contain more complex information compared with other sensors, it is worthwhile to investigate the influence of different map representations on the model performance. Here, two kinds of map representations are evaluated, the image-based and vector-based methods. Both extract information from an ASAM OpenDRIVE file, and the difference is the way to describe the information extracted. In the image-based representation, the map is rasterized from the BEV perspective using the lane and road messages in the file, and a map is drawn using the map elements in the raster map, as illustrated in Fig. \ref{fig:arc} (c). For the vector-based representation, the nearby center lines of the lanes in the map are vectorized. Here, a window of $28\times28$ m centered in the ego vehicle position is used to define the surrounding map elements, and the vectorized map is shown schematically in Fig. \ref{fig:arc} (d). What distinguishes these two map types from each other are the orientation and semantic information of the map elements, which will be discussed in detail in Section \ref{sec:local_route}.

\subsubsection{Radar}
\label{sec:radar_preprocess}
Endowed with the velocity information, radar contributes to the following and lane switching maneuvers, especially in highly dynamic environments. In our method, two radars with the specification of a 35$^\circ$ field of view (FOV) and a maximum range of 100 meters is arranged, one at the front end of the vehicle, and the other at the rear of the vehicle. To avoid radar waves being reflected from the surface of the ground, the pitch angle of the two radars is increased by 5$^\circ$, and their height is set as 1 m. We first calculate the time for each point to reach the radar sensor position by dividing the point depth by the velocity of this point, and then we select the points closer to the radar sensor according to the time calculated. In our setting, the top $N=81$ radar points are selected, and if the actual number of radar points is less than $N$, the remaining feature vectors are padded with 0.

\subsubsection{LiDAR Point Cloud}:
\label{section:pc}
Our approach to dealing with LiDAR data is similar to \cite{rhinehart2019precog}, which converts the 3D LiDAR data into a 2D BEV grid map by calculating the number of LiDAR points inside each grid as Fig. \ref{fig:arc} (a) and (b) show. The total area considered by the 2D BEV grid map is $32\times32$ m, with 28 m distance in front of the vehicle, 6 m distance behind the vehicle, and 16 m distance on the left and right sides of the vehicle. The reason for considering the LiDAR points at the rear of the agent is that the information in this space is crucial when lane changing happens or when the ego vehicle drives away from the highway. 
Other settings are the same as \cite{transfuser}.
% The resolution of the grid map is the same as that in \cite{transfuser}, which is set as $0.125\times0.125$ m. To separate the LiDAR points in terms of height, two channels are formed with one below 2 m high and the other one above 2 m high. Therefore, the number of the total pixels in the 2D pseudo image is $256\times256\times2$.

\subsubsection{Camera Images}
For the RGB input, one camera is deployed in front of the vehicle with a 100$^\circ$ FOV and a $400\times300$ resolution in pixels. Because of the distortion caused by the rendering of the cameras in the CARLA simulator, the RGB images are cropped to $256\times256\times3$. 

\subsubsection{Output Representation}:
As in \cite{transfuser} and \cite{neat}, the output of the network forecasts the future trajectory $\mathbf w$ of the vehicle in the BEV space which coincides with the ego vehicle's coordinates frame. The trajectory is represented by a series of 2D waypoints in the form of $\{\mathbf w_t = (x_t, y_t)^{T}_{t=1}\}$, where the default number of waypoints is set as 4.

\subsection{Architecture Design}
% 插入Vectornet的框架与描述
In this section, the network architecture of the proposed approach is discussed. To better fuse the network inputs with multiple modalities, different sensor data inputs are treated in different ways before the fusion layers, as indicated in Fig. \ref{fig:arc}. Thus the remainder of this section introduces the part of the network architecture before the fusion layers, and the fusion mechanism\cite{transfuser}.  
\begin{figure}[!t]
\vspace{-0.7em}
\begin{algorithm}[H]
\caption{OpenDrive to VectorNet Input}\label{alg:op2v}
\begin{algorithmic}[1]
  \State \textbf{Initialize}: $\mathcal{S}_{\text{rough\_lane}} \gets \emptyset$ \\
  $\qquad \quad \quad\; \mathcal{S}_{\text{lane}}\gets \textsc{Parse}(\text{opendrive})$
  \For{\textit{lane} in $\mathcal{S}_{\text{lane}}$}
  \State $\mathcal{L}_{\text{lane\_node}} \gets \emptyset$
        \For{\textit s in \{$0,\dots,\textit{lane}.max\_s$\}}
            \State $\textit{lane\_node} \gets \textsc{CalculateProperty}(\text{lane,s})$
            \State $\mathcal{L}_{\text{lane\_node}}\textsc{.append}(\textit{lane\_node})$
            
            \If{$\mathcal{L}_{\text{lane\_node}}\textsc{.size} \ge 10$}
                 \State $\mathcal{S}_{\text{rough\_lane}}\textsc{.insert}(\mathcal{L}_{\text{lane\_node}})$
                 \State $\mathcal{L}_{\text{lane\_node}} \gets \emptyset$
            \EndIf
        \EndFor
        \If{$\mathcal{L}_{\text{lane\_node}}  \ne \emptyset$}
            \State $\mathcal{S}_{\text{rough\_lane}}\textsc{.insert}(\mathcal{L}_{\text{lane\_node}})$
        \EndIf
  \EndFor
  \State \Return $\mathcal{S}_{\text{rough\_lane}}$
\end{algorithmic}
\end{algorithm}
\vspace{-2.5em}
\end{figure}
% discretize the lanes
\subsubsection{Map}
The ASAM OpenDRIVE file is parsed into a discrete HD map, and this process is summarized in Algorithm \ref{alg:op2v}, where $\mathcal{S}_{\text{lane}}$ refers to the set consisting of the \textit{lane}s expressed by analytic formulas, $\mathcal{L}_{\text{lane\_node}}$ corresponds to a list of discretized points of the \textit{lane}, and $\mathcal{S}_{\text{rough\_lane}}$ is the set composed of discretized lanes called \textit{rough lane}s and $\textit s$ is the road arc length from the start to the reference point.
To select the lanes of interest around the ego vehicle so as to discretize them, all lanes inside a window centered at the ego vehicle are selected. In our setting, the number of lanes is $N$, and each discretized lane is represented by $P=10$ points and $P-1$ vectors, with each vector expressed by:
\begin{align}
\mathbf{v}_{i}=\left[\mathbf{d}_{i-1}, \mathbf{d}_{i}, \mathbf{a}_{i}\right], i\in \left [1,\dots,P \right]
\end{align}
where $\mathbf{d}_{i-1}$, $\mathbf{d}_i$ are the coordinates of each lane point $(x,y)$, and $\mathbf a_i$ is the label of this lane point. In correspondence with the available semantic labels of a point in the OpenDRIVE message, here $\mathbf a_i$ could indicate whether the lane point is at a junction and if it is available for left/right change.

% vectorize the lanes
After discretizing the lanes selected as mentioned above, these lanes are then vectorized according to \cite{vectornet}. Briefly speaking, the lane vectors are sent to their own Multilayer Perceptron (MLP), followed by attention layers to concatenate the outputs of the MLPs. We adopt \textit{VectorNet} to process the lane features. The map feature information output from VectorNet is then fed to ResNet18, as demonstrated in Fig. \ref{fig:arc} to get the OpenDRIVE map data in the same size as the other sensor data before the fusion layers.
\subsubsection{Radar}
% architecture of radar data
When dealing with radar inputs, if directly resizing them to the same size as that of the camera data before fusion, significant amounts of information will be lost because of the sparsity of radar points. Inspired by \cite{GAT},  we choose to connect the radar points to form a graph whose weights are the relative azimuth distances. Then the weights of this graph are multiplied by the radar points to get the radar features as follows:
\begin{align}
\mathbf{F}^{out}=\mathbf{W} \mathbf{P}^{i n},
\end{align}
where $\mathbf F^{out} \in \mathbb{R}^{N \times L}$ is the output radar feature, $\mathbf P^{in} \in \mathbb{R}^{N \times L}$ is the input radar points, $\mathbf{W} \in \mathbb{R}^{N \times N}$ is the weight matrix obtained from the weights of the graph, $N=81$ is the number of radar points after pre-processing mentioned in Section \ref{sec:radar_preprocess}, and $L=5$ is the number of the feature labels, which include a point's velocity, depth, azimuth, altitude and the label to indicate its location. These radar features are sent to the attention layers to be resized to the same size as other sensor data. The overall processing procedure for radar data mentioned above is the Graph Attention Network shown in Fig. \ref{fig:arc}.
\subsubsection{Fusion}

\begin{figure}[t!]
\centering
\vspace{2em}
\includegraphics[width=3.4in]{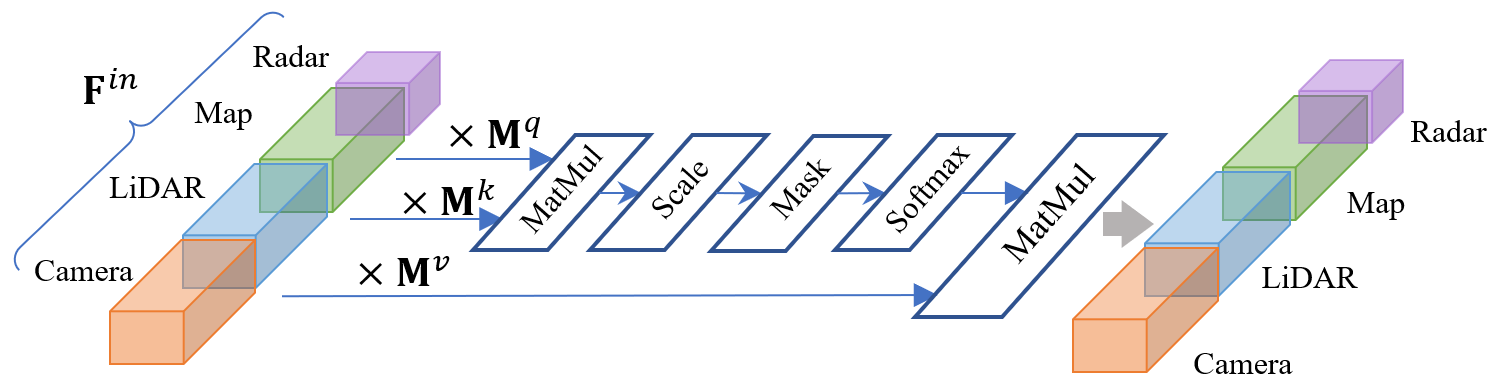}
\caption{Detailed version of the gray fusion layer block in Fig. \ref{fig:arc}}
\label{fig:fusion}
\vspace{-1.0em} % for figure gap smaller
\end{figure}

With different sensor features of the same size, transform layers\cite{transfuser} are utilized to offer a chance for these sensor data to `communicate' with each other. There are four fusion layers in total, with each receiving the concatenated multi-modal sensor features extracted at different stages of their own feature networks. After being blended with other types of sensor data in fusion layers, these features are then sent back to the place from which they are extracted before fusion, as shown in Fig. \ref{fig:arc}. The detailed network architecture inside the fusion layers is demonstrated in Fig. \ref{fig:fusion}. It shows that all the different types of sensor data are concatenated together as the input of the fusion layers $\mathbf{F}^{i n}$, which is then multiplied by three matrices: 
\begin{align}
\mathbf{Q}=\mathbf{F}^{i n} \mathbf{M}^{q}, \mathbf{K}=\mathbf{F}^{i n} \mathbf{M}^{k}, \mathbf{V}=\mathbf{F}^{i n} \mathbf{M}^{v},
\end{align}
where $\mathbf {Q}$, $\mathbf{K}$ and $\mathbf{V}$ refer to the query, key and value respectively, and $\mathbf{M}^{q} \in \mathbb{R}^{D_{f} \times D_{q}}, \mathbf{M}^{k} \in \mathbb{R}^{D_{f} \times D_{k}},\mathbf{M}^{v} \in \mathbb{R}^{D_{f} \times D_{v}}$ are weight matrices in the form of linear layers.

The overall process of Attention mechanism\cite{vaswani2017attention} in Fig. \ref{fig:fusion} could be generalized as: 
\begin{align}
\mathbf{A}=\operatorname{softmax}\left(\frac{\mathbf{Q K}^{T}}{\sqrt{D_{k}}}\right) \mathbf{V},
\label{fusion_eq}
\end{align}

% output of the network
\subsubsection{Output}
The output layers of the overall network in Fig. \ref{fig:arc} are the same as those in \cite{transfuser} and \cite{neat}, which uses four cascaded Gated Recurrent Units (GRUs) for the final waypoint output. The first GRU takes the differential ego vehicle waypoints and the relative goal location points as inputs, and the last GRU outputs the coordinates of the predicted waypoints relative to the ego vehicle. The loss function of the network is set as the L1 loss between the expert's and the agent's waypoints:
\begin{align}
\mathcal{L}=\sum_{t=1}^{T}\left\|\mathbf{w}_{t}-\mathbf{w}_{t}^{g t}\right\|_{1},
\end{align}
where $\mathbf{w}_{t}^{g t}$ is expert waypoint coordinates on time $t$.
% output of the whole network to car control
After training, the network can receive the sensor data and relative goal location to output the predicted waypoints. Two PID controllers are connected with the network to transform the network outputs to the control signals on the steering wheel, throttle, and brake pedal.

\section{EXPERIMENTS}
\label{sec:experiment}
Since it is costly to collect large-scale multi-modality sensor data in the real world, the CARLA simulator \cite{carla} is chosen for generating the data needed in the simulation environment. Moreover, it also provides public routes and scenarios for users to evaluate their own agents based on the unified criteria in the online leaderboard platform. To maintain fairness, the AUTO\cite{lbc} and MMFN experts collect the same routes for training the models in Table \ref{table_general} which have 207K frames data separately.

% Our training set includes all Town maps from tiny to short routes defined by \cite{transfuser}. Town05 Long sets as testing routes during training. Training scenarios are provided by leaderboard who public three available scenarios for training agents which include: control loss due to bad conditions, obstacle avoidance in an emergency, or yield for pedestrians.
% different test dataset and source of pre-trained models

In Section \ref{sec:local_route}, to show the generalization performance of the trained model, instead of using a single map in one Town, the routes in \cite{neat} are used for testing instead. Every two routes out of these twelve routes are obtained from the same Town and of the same length. Considering that the main purpose is to test whether the agent can handle complex scenarios, we add all the events under the available driving scenarios in the CARLA leaderboard into the Town maps.
% overall intro to leaderboard
\subsection{Evaluation Metric}
The \href{https://leaderboard.CARLA.org/#evaluation-and-metrics}{CARLA leaderboard} includes unified metrics to evaluate the driving task, the driving score (DS), route completion (RC), and infraction penalty, among which the most important, the DS, is computed by $R_i \times P_i$, where $R_i$ and $P_i$ refer to the RC and infraction penalty coefficient of the $i$-th route respectively.

The DS has an upper limit of 100, and a higher DS indicates a better agent. The RC reveals several facts, e.g., whether the agent can successfully go through junctions without traffic lights. The infraction results shown in Table \ref{expert} and \ref{table_general} are obtained by calculating the infractions of the vehicle per kilometer (Infra/km). Thus, the lower the Infra/km, the better the agent behaves. 
% To be more specific, the infractions include collisions with layout, pedestrians, and vehicles, deviating from the route lane, and breaking the rules of traffic lights. 
More details about infractions can be obtained from \href{https://leaderboard.carla.org/#evaluation-and-metrics}{CARLA leaderboard}.

\subsection{Expert Performance}
\label{sec:expert_evaluation}
% performance of expert
As seen from Table \ref{expert}, our MMFN expert outperforms the AUTO expert in most cases. The reason is mainly that the MMFN expert abides by more reasonable rules similar to those in real driving scenarios. Especially for Town 03, 04, and 05, where there are more complex scenarios like lane changing on highways, the MMFN expert performs considerably better than the AUTO expert. Overall, the infractions per kilometer decreased by about 45.27\% for all routes in the six towns, indicating that our expert has more awareness of safe driving. Furthermore, the next section also shows that as the upper bound of how well the agent can behave, the expert plays an important role in the driving performance of its agent. 
\renewcommand{\arraystretch}{1.4}
\begin{table}[t!]
\caption{Expert Driving Performance}
\begin{center}
\begin{tabular}{c|ll|ll|ll} 
\Xhline{2\arrayrulewidth}
\multicolumn{1}{c}{\multirow{2}{*}{\begin{tabular}[c]{@{}c@{}}No. \\Town\end{tabular}}} & \multicolumn{2}{c}{DS $\uparrow$}          & \multicolumn{2}{c}{RC $\uparrow$}          & \multicolumn{2}{c}{Infra/km $\downarrow$}  \\ 
\cline{2-7}
\multicolumn{1}{c}{}    & AUTO           & MMFN           & AUTO           & MMFN           & AUTO & MMFN                \\ 
\hline
1                       & 85.90          & \textbf{94.00}  & \textbf{100.00}             & \textbf{100.00}   & 0.06 & \textbf{0.02}       \\
2                       & \textbf{62.58} & 60.06           & \textbf{95.29}     & 85.09             & 0.16 & \textbf{0.12}       \\
3                       & 66.36          & \textbf{79.55}  & 80.91              & \textbf{88.07}    & 0.07 & \textbf{0.04}       \\
4                       & 88.79          & \textbf{91.90}  & 98.36              & \textbf{98.7}     & \textbf{0.01} & \textbf{0.01}       \\
5                       & 71.00          & \textbf{87.60}  & \textbf{100.00}    & 94.6              & 0.07 & \textbf{0.02}       \\
6                       & 60.92          & \textbf{89.60}  & \textbf{100.00}    & \textbf{100.00}   & 0.06 & \textbf{0.01}       \\

\Xhline{2\arrayrulewidth}
\end{tabular}
\label{expert}
\end{center}
\vspace{-1.5em} % for figure gap smaller
\end{table}
\renewcommand{\arraystretch}{1.0}

\subsection{Local Evaluation}
\label{sec:local_route}
\renewcommand{\arraystretch}{1.4}
\begin{table*}[ht!] %ht
\caption{Driving performance of different models with different expert data}
\centering
\begin{tabular}{llll|lll} 
\Xhline{2\arrayrulewidth}
                                         & \multicolumn{3}{c}{AUTO expert}                  & \multicolumn{3}{c}{MMFN expert}                    \\ 
\cline{2-7}
Methods                                  & Driving Score $\uparrow$ & Route Completion $\uparrow$ & Infra/km $\downarrow$ & Driving Score $\uparrow$ & Route Completion $\uparrow$ & Infra/km  $\downarrow$  \\ 
\hline
CILRS\cite{cilrs}           & 20.52 $\pm$ 1.07      & 33.05 $\pm$ 0.08         & 1.70 $\pm$ 0.01           & 18.18 $\pm$ 3.44              & 30.04 $\pm$ 4.17         & 1.65 $\pm$ 0.05   \\
AIM\cite{transfuser}              & 67.56 $\pm$ 1.77      & 86.53 $\pm$ 5.14         & 1.03 $\pm$ 0.02           & 73.45 $\pm$ 1.04              & 90.33 $\pm$ 0.12         & 0.73 $\pm$ 0.02   \\
Transfuser\cite{transfuser} & 56.25 $\pm$ 0.24      & 62.66 $\pm$ 0.34         & 0.64 $\pm$ 0.17           & 62.36 $\pm$ 0.58              & 79.43 $\pm$ 0.29         & 0.97 $\pm$ 0.20   \\
MMFN (Radar)                & 46.05 $\pm$ 5.95      & 58.82 $\pm$ 8.73         & 0.91 $\pm$ 0.08           & 58.72 $\pm$ 2.47              & 68.25 $\pm$ 3.96         & 0.67 $\pm$ 0.08   \\
MMFN (Image)                & 72.22 $\pm$ 3.51      & \textbf{88.78} $\pm$ 0.43         & 0.70 $\pm$ 0.11           & 76.56 $\pm$ 0.41              & 89.51 $\pm$ 1.09         & 0.67 $\pm$ 0.08   \\
MMFN (VectorNet)            & \textbf{75.62} $\pm$ 0.32      & 82.17 $\pm$ 1.73         & \textbf{0.50} $\pm$ 0.08           & \textbf{88.75} $\pm$ 1.42     & \textbf{96.53} $\pm$ 0.47    & \textbf{0.41} $\pm$ 0.05   \\ 
\hline
\rowcolor[rgb]{0.906,0.902,0.902} Expert & 84.08 $\pm$ 1.67      & 98.62 $\pm$ 0.23         & 0.35 $\pm$ 0.04  & 92.32 $\pm$ 1.52      & 97.32 $\pm$ 0.25         & 0.17 $\pm$ 0.02   \\
\Xhline{2\arrayrulewidth}
\end{tabular}
\label{table_general}
\end{table*}
\renewcommand{\arraystretch}{1.0}

% result analysis
From Table \ref{table_general}, we can conclude that our MMFN expert can improve the agent's driving performance by comparing the left and right columns. The DS and RC increase by about 9.8\% and 10.3\% respectively, and the infra/km decreases by around 4.9\%. Moreover, comparing the left and right columns of one row in Table \ref{table_general} reveals that the expert performance also affects the agent performance, since the expert limits how well the agent can behave, as mentioned in Section \ref{sec:expert_evaluation}. 

Before row-wise comparisons, the three baselines given in Table \ref{table_general} need to be introduced. CILRS\cite{cilrs} and AIM\cite{transfuser} both use one camera as the network input, which is then sent to a ResNet 34. CILRS outputs the vehicle control signal directly, while AIM outputs waypoints through four GRU decoders followed by PID controllers. Based on AIM, Transfuer\cite{transfuser} adds the LiDAR input and transform layers to fuse the camera and LiDAR data. Apart from the camera and LiDAR, MMFN adds two more sensor data, HD map, and radar, into the network to offer more information. MMFN (Image) transforms the HD map into a BEV image which is then fed into ResNet 18. By contrast, MMFN (VectorNet) vectorizes the HD map first and extracts features from the vectorized map. Additionally, MMFN (Radar) adds radar input based on MMFN (VectorNet).

%, not all sensors are beneficial to results & HD map is good for driving performance 
It experimentally shows that how to represent the newly added sensor data is crucial in improving the driving performance, by comparing the results of Transfuser and AIM in Table \ref{table_general}. Although one LiDAR and one camera are used in Transfuser, the driving performance of this model is still worse than that of AIM which only uses one camera. The results of MMFN (Image), MMFN (VectorNet), and Transfuser in Table \ref{table_general} proves the effective use of the additional HD map data, either in the image form or in the vector form, since the remaining sensors except the map used in MMFN are the same as those used in Transfuser. 
% 给一个具体的场景结果图 -> TODO 
\begin{figure}[!t]
\centering
\vspace{0.5em}
\includegraphics[width=3.0in]{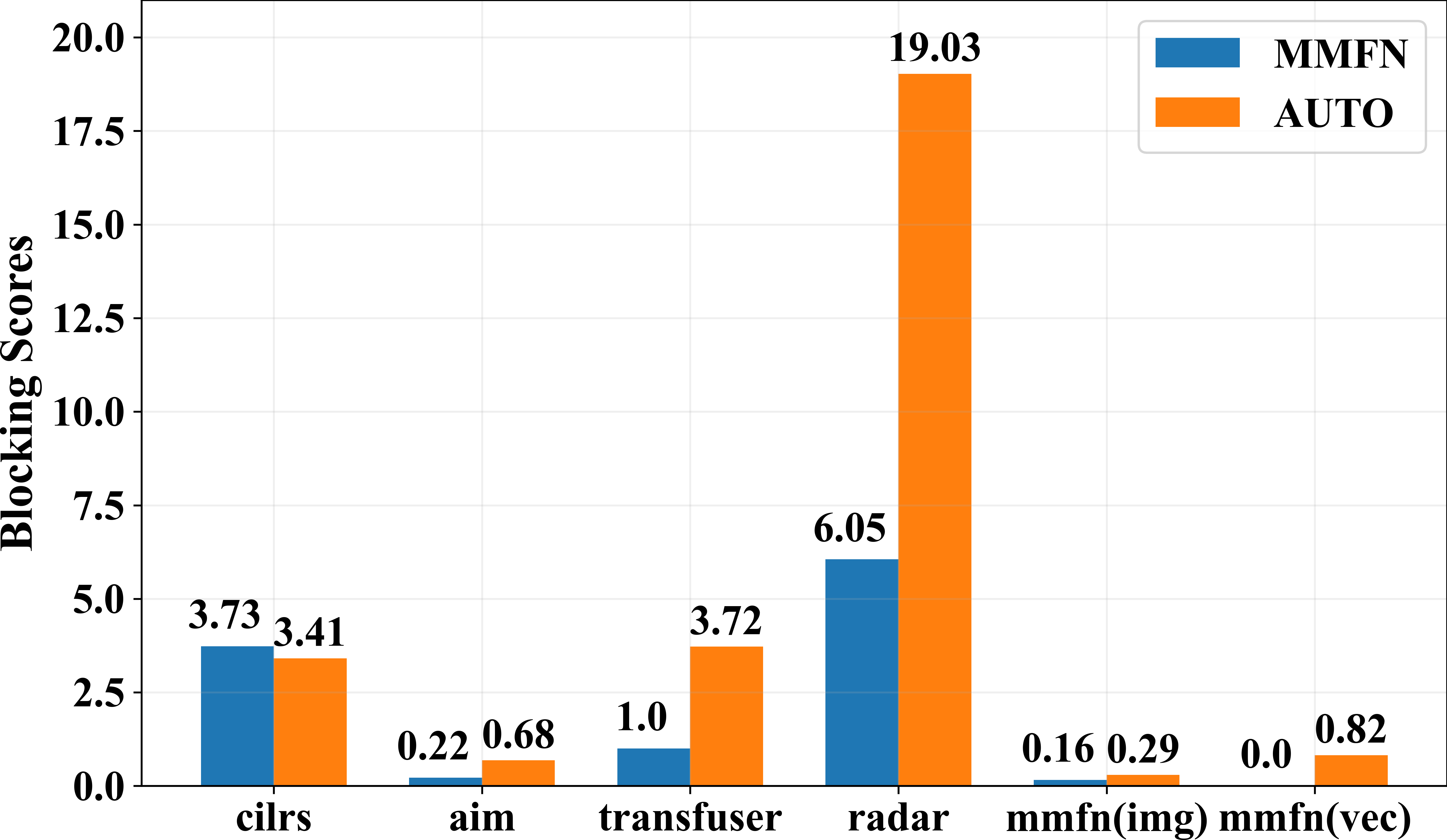}
\caption{Blocking scores of each model with different experts on local routes, with lower blocking scores equivalent to higher route completion}
\label{fig:blocking}
\vspace{-1.0em} % for figure gap smaller
\end{figure}

% VectorNet vs. image map
Analyzing the results of MMFN (Image) and MMFN (VectorNet) in Table \ref{table_general} further shows that vectorizing the HD map is superior to using images to represent the HD map which can be seen as one of the ablation studies.
Possible reasons for it are twofold. One is that convolutional neural networks (CNNs) used in MMFN (Image) are not originally designed to capture long-range geometry structures as discussed in \cite{vectornet}, while VectorNet can extract more map element features instead. The other one is that the map elements in VectorNet offer some information more directly like orientation or semantic labels, so the agent can give more accurate commands with a clearer understanding of the surroundings.

%reasons for the bad performance of radar
In Table \ref{table_general}, MMFN (Radar) fails to compete with MMFN (Image), MMFN (VectorNet), and even with Transfuser \cite{transfuser}, though it has one more radar sensor compared with the other 3 models. It is demonstrated in Fig. \ref{fig:blocking} that blocking frequently happens in MMFN (Radar), which means the additional use of radar makes the agent more alert to potentially dangerous driving scenarios. One possible reason for the unsatisfactory performance may be that the embeddings of radar data are not good enough to clearly interpret the messages from the raw radar points, and thus it may mislead other types of sensor data in fusion layers.

\subsection{Online Leaderboard}
\renewcommand{\arraystretch}{1.4}
\begin{table}
\caption{Online Leaderboard Results}
\centering
\begin{tabular}{lllll}
\Xhline{2\arrayrulewidth}
Methods                                        & Sensors             & DS $\uparrow$    & RC $\uparrow$   & $P$ $\uparrow$    \\ 
\hline
\rowcolor[rgb]{0.929,0.929,0.929} MMFN$^\star$ (Ours) & 1 Cameras + 1 LiDAR & \textbf{22.8}  & 47.22         & 0.63  \\
NEAT\cite{neat}                                       & 3 Cameras           & 21.83          & 41.71         & \textbf{0.65}  \\
AIM-MT\cite{neat}                                     & 1 Cameras           & 19.38          & 67.02         & 0.39  \\
T4AC+$^\star$ \cite{robesafe_ads}                     & 1 Cameras + 1 LiDAR & 18.75          & \textbf{75.11}& 0.28  \\
TransFuser\cite{transfuser}                           & 1 Cameras + 1 LiDAR & 16.93          & 51.82         & 0.42  \\
Pylot$^\star$\cite{gog2021pylot}                      & 2 Cameras + 1 LiDAR & 16.7           & 48.63         & 0.5   \\
CaRINA$^\star$\cite{carina}                           & 2 Cameras + 1 LiDAR & 15.55          & 40.63         & 0.47  \\
\Xhline{2\arrayrulewidth}
\multicolumn{4}{l}{\small $^\star$ means the agent use the HD Map}
\end{tabular}
\label{table_online}
\vspace{-1.5em}
\end{table}
\renewcommand{\arraystretch}{1.0}

% intro to the online leaderboard
There are two tracks in the CARLA online leaderboard, the sensor track, and the map track. Since the HD map is used in this method, in Table \ref{table_online}, we mainly compare our model with several top models in the map track, but a few models in the sensor track are also evaluated for a more comprehensive comparison. Unlike the experiments in Section \ref{sec:local_route}, all the routes and scenarios in the online leaderboard are not public for the sake of fairness. The models are ordered by DS as it is the most important metric to evaluate a model. Another metric $P$ in this table is the infraction penalty defined by the leaderboard, which is initialized as 1 and will be subtracted by different infraction penalties during evaluation. The results in Table \ref{table_online} are all copied from the online leaderboard before the date 25/02/2021.

% result analysis
In the online leaderboard, among all the models which use the HD map data in the map track, our model shows the best performance. Compared with other models using multiple sensors in the sensor track, our model has a higher infraction penalty coefficient represented by $P$, especially even with relatively low RC. Our agent intends to behave more conservatively in complicated driving scenarios, resulting in fewer infractions.

\section{CONCLUSION}
In conclusion, our proposed approach shows the effective use of the additional HD map data in the end-to-end driving task. It also proves that using \textit{VectorNet} to represent this data achieves superior driving performance to using a BEV raster image. We explored how to represent all the sensor data and proposed a Multi-Modal-Fusion-Network (MMFN) to use camera images, LiDAR point clouds, an OpenDRIVE map, and radar as the network inputs for the end-to-end autonomous driving. Furthermore, the expert presented in this work also contributes to the agent's performance. 

We hope that this work may arouse the interest in the research community about using HD maps, radar, and other sensors in end-to-end driving. The performance of our model could be further enhanced by setting up more rules for the expert to abide by. As suggested in \cite{GRI}, \cite{marl} and \cite{liang2019multi}, considering more maneuvers of the expert or replacing our agent with a costly deep reinforcement learning agent\cite{zhang2021end} could be helpful in improving the performance. Our future work will also include designing sub-tasks, like classification and segmentation using cameras or LiDAR, before the output layers to help the agent to learn the final driving task more quickly and accurately. 
% \clearpage
\addtolength{\textheight}{-12cm}   % This command serves to balance the column lengths



\section*{ACKNOWLEDGMENT}
Thanks to HKUST Ramlab's members: Jin Wu, Jie Cheng, Bowen Yang, Xiaodong Mei, and Peide Cai who gave constructive comments on this work. We also thank the anonymous reviewers for their constructive comments.
This work was supported by Guangdong Basic and Applied Basic Research Foundation, under project 2021B1515120032,  awarded to Prof. Lujia Wang.
% \clearpage
%%%%%%%%%%%%%%%%%%%%%%%%%%%%%%%%%%%%%%%%%%%%%%%%%%%%%%%%%%%%%%%%%%%%%%%%%%%%%%%%
% ================================= reference ================================ %
% \vspace{30\baselineskip}% Try 29\baselineskip instead!
\bibliographystyle{IEEEtran}
\bibliography{IEEEabrv,ref}

\end{document}